\documentclass{article}

\usepackage{arxiv}

\usepackage[utf8]{inputenc} 
\usepackage[T1]{fontenc}    
\usepackage{hyperref}       
\usepackage{url}            
\usepackage{booktabs}       
\usepackage{amsfonts}       
\usepackage{nicefrac}       
\usepackage{microtype}      
\usepackage{lipsum}		
\usepackage{graphicx}
\usepackage{natbib}
\usepackage{doi}

\usepackage{amsmath}
\usepackage{multirow}
\usepackage{caption}
\usepackage{subcaption}
\usepackage{xcolor}

\title{VICTOR: Visual Incompatibility Detection with Transformers and Fashion-specific contrastive pre-training}

\author{ 
    Stefanos-Iordanis Papadopoulos \\
	CERTH-ITI\\
	\texttt{stefpapad@iti.gr} \\
	\and
	Christos Koutlis \\ 
	CERTH-ITI \\
	ckoutlis@iti.gr\\
	\and
	Symeon Papadopoulos\\
	CERTH-ITI\\
	papadop@iti.gr\\
    \and
    Ioannis Kompatsiaris\\
    CERTH-ITI\\
    ikom@iti.gr\\
}

\date{}

\renewcommand{\shorttitle}{VICTOR: Visual Incompatibility Detection}

\hypersetup{
pdftitle={A template for the arxiv style},
pdfsubject={q-bio.NC, q-bio.QM},
pdfauthor={David S.~Hippocampus, Elias D.~Striatum},
pdfkeywords={First keyword, Second keyword, More},
}

\begin{document}
\maketitle

\begin{abstract}
For fashion outfits to be considered aesthetically pleasing, the garments that constitute them need to be compatible in terms of visual aspects, such as style, category and color. Previous works have defined visual compatibility as a binary classification task with items in a garment being considered as fully compatible or fully incompatible. However, this is not applicable to Outfit Maker applications where users create their own outfits and need to know which specific items may be incompatible with the rest of the outfit.  To address this, we propose the Visual InCompatibility TransfORmer (VICTOR) that is optimized for two tasks: 1) overall compatibility as regression and 2) the detection of mismatching items and utilize fashion-specific contrastive language-image pre-training for fine tuning computer vision neural networks on fashion imagery. We build upon the Polyvore outfit benchmark to generate partially mismatching outfits, creating a new dataset termed Polyvore-MISFITs, that is used to train VICTOR. 
A series of ablation and comparative analyses show that the proposed architecture can compete and even surpass the current state-of-the-art on Polyvore datasets while reducing the instance-wise floating operations by 88\%, striking a balance between high performance and efficiency. We release our code at \url{https://github.com/stevejpapad/Visual-InCompatibility-Transformer}
\end{abstract}


\section{Introduction}
\label{sec:intro}

Fashion products do not exist in a vacuum. When customers consider buying a new garment they may contemplate its subjective appeal, price, quality or trendiness but also think of ways to match it with other pieces and how compatible it is with other items in their wardrobe.
To help customers in their endeavours, contemporary e-commerce applications usually provide outfit recommendations and suggestions of how to ``complete the look'' based on an item of interest.
Outfit compatibility is a rather challenging task: not only is it highly subjective but it also involves numerous variables such as the style, color, fit, patterns,
proportions, textures of numerous garments and how these aspects interrelate.
To this end, researchers have recently utilized computer vision neural networks, that learn to produce informative representations from fashion images, along with pairwise-based \cite{tan2019learning, vasileva2018learning}, graph-based \cite{cucurull2019context, cui2019dressing} or attention-based neural networks \cite{zhan2021pan, zhan20213, chen2019pog} that learn to predict the compatibility of outfits. 

However, previous studies define outfit compatibility prediction as a binary ($OC_{b}$) classification task. An outfit is either fully compatible or fully incompatible. 
This is a reasonable assumption for e-commerce applications that recommend fully compatible outfits to their customers.
It is not as applicable to \textit{Outfit Maker} applications\footnote{Examples of outfit maker applications include: ShopLook, Smart Closet, Stylebook, Pureple and Combyne}, where users combine garments to create their own outfits. 
Instead, it would be more useful to offer an overall compatibility score and detect specific mismatching garments in order to inform users which items are not compatible with the rest of the outfit.
This would give a sense of how aesthetically pleasing an outfit is and help users identify garments with clashing colors or patterns, select more suitable alternatives and generally fine-tune their outfits.

In this study we define outfit compatibility as a regression ($OC_{r}$) problem and also address the task of mismatching item detection (MID) in fashion outfits.
To the best of the authors' knowledge, no previous study has addressed these tasks.
We use the Polyvore outfit dataset \cite{vasileva2018learning} which consists of fully compatible and incompatible outfits to generate partially mismatching outfits (MISFITs).
We propose the \textbf{V}isual \textbf{I}n\textbf{C}ompatibility \textbf{T}ransf\textbf{OR}mer, or VICTOR, a multi-tasking, Transformer-based architecture that is trained to predict the overall $OC_{r}$ score and detect mismatching garments in an outfit.
Previous works on $OC_{b}$ either rely on feature extraction from computer vision models pre-trained on ImageNet\cite{chen2019pog, lorbert2021scalable} or end-to-end fine-tuning \cite{han2017learning, vasileva2018learning, lin2020fashion, tan2019learning, sarkar2022outfittransformer}.
While E2E fine-tuning tends to significantly outperform feature extraction, it is notably more resource intensive. Instead, we utilize \textit{fashion-specific contrastive language image pre-training} (FLIP) to fine-tune computer vision models for fashion imagery and then use the extracted visual features for $OC_{r}$ and MID.
The ablation study showed that multi-tasking outperforms the single-tasking and that multi-modality improves upon the visual-only versions on VICTOR while the comparative analysis showed that VICTOR with FLIP are capable of competing and even surpassing, the current state-of-the-art on Polyvore datasets for $OC_{b}$ while reducing instance-wise floating point operations (FLOPs) by an impressive 88\%.

The main contributions of our work are:

\begin{itemize}

\item We define two new sub-tasks around visual compatibility, namely: outfit compatibility prediction as regression ($OC_{r}$) and mismatching item detection (MID) and examine them in the domain of Fashion.

\item We propose VICTOR, a multi-tasking Transformer-based neural network that is optimized for both tasks and 
utilize fashion-specific contrastive language image pre-training (FLIP) for fine-tuning computer vision neural networks on fashion imagery. 

\item We experiment with four computer vision backbone networks and perform an extensive ablation and comparative analysis that shows VICTOR with FLIP to be capable of competing and even surpassing the current state-of-the-art on Polyvore datasets while reducing instance-wise floating point operations by 88\% and total study-wise operations by up to 98\%.

\end{itemize}

\section{Related Work}
\label{sec:rw}

In recent years, researchers have shown increased interest in applying deep learning and computer vision neural networks \cite{cheng2021fashion} in order to address numerous tasks relevant for the Fashion domain including category and attribute classification \cite{liu2016deepfashion, papadopoulos2022attentive}, trend forecasting \cite{al2017fashion, mall2019geostyle}, popularity prediction \cite{skenderi2021well, papadopoulos2022multimodal}, fashion recommendations systems \cite{hwangbo2018recommendation, stefani2019cfrs} and among them, the task of outfit recommendations. In order to recommend complete outfits it is first necessary to understand which garments go well together and can create compatible and cohesive outfits. 

The first studies to address the task, considered outfit compatibility as a series of pairwise comparisons between all comprising garments \cite{tan2019learning, vasileva2018learning}. Pairwise-based approaches have utilized Siamese \cite{veit2015learning} and triplet loss networks with either type-aware embeddings \cite{vasileva2018learning} or similarity-aware embeddings \cite{tan2019learning}. 
Other works, instead of aggregating garment-level relations
attempted to capture global outfit-level representations with the use of bidirectional LSTMs \cite{han2017learning} or graph neural networks \cite{cucurull2019context, cui2019dressing}. In practice, outfits are not ordered sequences; the order of the garments should not affect the model's predictions. Thus, recurrent neural networks are not the most suitable architecture for the task. 
On the other hand, graph-based approaches tend to require large ``neighborhoods'' of compatible garment-nodes
as input in order to reach optimal performance which is problematic for new items that lack neighbor information and may straggle from the cold start problem \cite{lin2020fashion}. 

In order to address the aforementioned challenges, more recent works have employed attention-based methods \cite{zhan2021pan, zhan20213, chen2019pog}.
Attention mechanisms have been used in pairwise-based approaches \cite{lin2020fashion, taraviya2021personalized} 
but the Transformer architecture has been successfully used for personalised outfit recommendations \cite{chen2019pog} and complementary item retrieval \cite{sarkar2022outfittransformer}. 
With the use of multi-head attention, the Transformer is suitable for learning relations between multiple items, in this case the compatibility between all garments in an outfit. Additionally, by removing the positional encoding \cite{vaswani2017attention, dosovitskiy2020image} it can capture unordered relations between all garments.

However, all aforementioned studies have defined outfit compatibility as a binary classification problem. An outfit is treated as either fully compatible or fully incompatible. To the best of our knowledge, this is the first study to tackle the task of mismatching item detection (MID) and treat compatibility prediction as a regression ($OC_r$) instead of a binary task ($OC_b$). 

Previous works have relied on visual, textual information and fashion categories for creating representations of garments in outfits. Transfer learning is generally being used for extracting visual information from the garment's images, either with feature extraction (FX) from ImageNet pretrained models \cite{chen2019pog, lorbert2021scalable} or by end-to-end fine-tuning (E2E) for $OC_b$ \cite{han2017learning, vasileva2018learning, lin2020fashion, tan2019learning, sarkar2022outfittransformer}. E2E tends to outperform FX-ImageNet since the visual features are trained to specialize on the target domain and task. Nevertheless, E2E is a highly resource intensive process since the gradients of a - usually large - network backbone need to be updated on top of the outfit matching neural network.
In this study, we attempt to find the middle ground between the efficiency of FX and the high accuracy of E2E by utilizing contrastive language-image pre-training - inspired by \cite{radford2021learning} - with a focus on fashion imagery. 

\section{Methodology}
\label{sec:methodology}

\subsection{Problem Formulation}

In this study, we address the task of mismatching item detection (MID) in fashion outfits. Moreover, we define visual outfit compatibility prediction as a regression task ($OC_r$) - allowing for partially mismatching outfits - in contrast to previous studies that define it as a binary classification task ($OC_b$).
Let a fashion outfit $\mathcal{O} = \{g_1, g_2, \dots, g_{n}\}$ consist of $n$ garments $g_i$. Our architecture after processing the outfit images, $\mathcal{I} = \{I(g_{1}), I(g_{2}), \dots, I(g_{n})\}$, produces $n+1$ outputs, one for the $OC_{r}$ task denoted $Y_{OC_r}\in (0,1) \subset \mathbb{R}$ and $n$ for the MID task denoted $Y_{MID}\in (0,1)^n \subset \mathbb{R}^n$, which are optimized to comply with the corresponding target variables, $T_{OC_r}$ and $T_{MID}$. First, $T_{OC_r} \in (0,1) \subset \mathbb{R}$ denotes the compatibility of the garments, where 0 means that all garments are incompatible, 1 that all are compatible and in-between values denote partial compatibility. Second, a list of binary values $T_{MID} = [x_{g_{1}}, x_{g_{2}}, \dots, x_{g_{n}}]$, with $x_{g_i} \in \{0,1\}$ and $i=1,\dots,n$, where 1 denotes the mismatching garments in outfit $\mathcal{O}$.
$OC_r$ is defined as a regression task and MID as a multi-label classification task. 

\subsection{Generating Mismatching Outfits}
\label{sec:gen}

Existing outfit datasets, e.g. Polyvore \cite{vasileva2018learning}, provide annotations for fully compatible or fully incompatible outfits. 
In this study we attempt to address partial incompatibility and the detection of specific mismatching items within an outfit. To this end, we generate partially mismatching outfits (MISFITs) with the following method. 
For every matching outfit $\mathcal{O}$, with $n>2$ we generate $m$ number of MISFITs by randomly selecting (i) the number of garments $1 \leq r \leq n-2$ that will be replaced and (ii) their positions $\mathcal{P}$. 
The garments in positions $\mathcal{P}$ are then replaced with randomly selected items of the same category, thus generating hard negative samples. Hard negative sampling forces the model to recognize and focus on fine-grained characteristics of the garments and their interrelations. In contrast, sampling garments from different categories would be easier for the model to recognize but it would learn less useful relations \cite{vasileva2018learning}, for example that an outfit can not consist of two dresses or two pairs of shoes. For $\mathcal{O}$ with $n = 3$ we only allow $r = 1$ because having outfits with only 1 compatible item is invalid. The target compatibility score is calculated as $T_{OC_r} = 1 - r/n$ and the mismatching items target is defined as a list $T_{MID}$ of binary values with 1 in $\mathcal{P}$ positions denoting the incompatible garments and 0 in other positions denoting the compatible garments. 
The fully compatible outfits retain $T_{OC_r}=1$ and $T_{MID} = [0, 0, \ldots,0]$, while the fully incompatible ones $T_{OC_r}=0$ and $T_{MID} = [1, 1, \ldots, 1]$, respectively.

\subsection{VICTOR}

\begin{figure*}[!ht]
    \centering
    \includegraphics[width=1\textwidth]{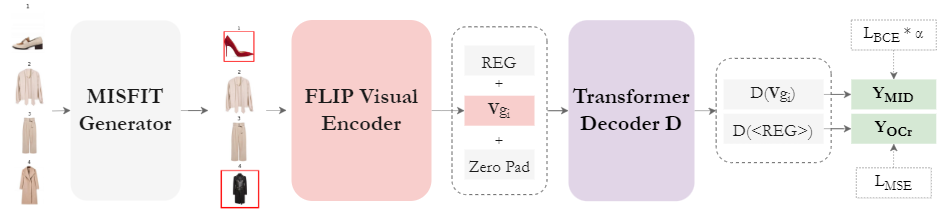}
    \caption{Workflow of the VICTOR architecture.}
    \label{fig:MISFITransformer}
\end{figure*}

The proposed pipeline of the Visual InCompatibility TransfORmer (VICTOR) is illustrated in Fig. \ref{fig:MISFITransformer}. 
First, the images $\mathcal{I} = \{I(g_{1}), I(g_{2}), \dots, I(g_{n})\}$ of all garments in an outfit $\mathcal{O}$ are passed through a visual encoder E\textsubscript{V}($\cdot$) that produces the corresponding vector representations $\mathbf{v}_{g_i}\in\mathbb{R}^{e\times 1}$, where $e$ is the encoder's embedding dimension. 
Then, following \citet{dosovitskiy2020image} that makes use of a classification token (CLS) we similarly consider a regression token (REG), a trainable vector that learns a global representation incorporating information about the relations of all garments in an outfit, and pass $\{\mathbf{v}_{g_i}\}_{i=1}^n\cup \{<\text{REG}>\}$ through a Transformer decoder\footnote{D($\cdot$) is actually structured as the encoder part of the original Transformer architecture but we use it to decode the image embeddings in our model thus we call it a decoder herein.} D($\cdot$).
Outfits are not sequential objects thus we do not make use of positional encodings \cite{vaswani2017attention, dosovitskiy2020image} so as to capture the unordered relations between garments.
The Transformer decoder D($\cdot$) consists of $L$ layers that have $h$ attention heads of embedding dimension $d$. Finally, the $OC_r$ score $Y_{OC_{r}}$ and the MID scores list $Y_{MID}$ of the outfit are calculated as:

\begin{equation}
\mathbf{v}_{g_i}=\text{E\textsubscript{V}}(I(g_i))
\end{equation}

\begin{equation}
\mathbf{d}_{g_i}=\text{D}(\mathbf{v}_{g_i})
\end{equation}

\begin{equation}
\mathbf{d}_{<\text{REG}>}=\text{D}(<\text{REG}>)
\end{equation}

\begin{equation}
Y_{OC_{r}}=\textbf{W}_1\cdot \text{GELU}(\textbf{W}_0\cdot \text{LN}(\mathbf{d}_{<\text{REG}>}))
\end{equation}

\begin{equation}
Y_{MID}[i]=\textbf{W}_i\cdot \text{GELU}(\text{LN}(\mathbf{d}_{g_i}))
\end{equation}

where $\mathbf{d}_{g_i}\in\mathbb{R}^{d\times 1}$ and $\mathbf{d}_{<\text{REG}>}\in\mathbb{R}^{d\times 1}$ are the Transformer's outputs, $\textbf{W}_0\in\mathbb{R}^{\frac{e}{2}\times d}$, $\textbf{W}_1\in\mathbb{R}^{1\times\frac{e}{2}}$ and $\textbf{W}_i\in\mathbb{R}^{1\times d}$ are sigmoid activated dense projection layers (learnable bias terms are considered but omitted here for clarity), LN stands for Layer Normalization and GELU is the activation function. Zero padding is also considered for D($\cdot$) input in outfits with less than 19 items being the largest outfit size in Polyvore. 

D($\cdot$) utilizes multi-head attention, thus each token contains information about a garment's interrelations with all other garments. In our case this translates to an item being mismatching with the rest of the items in the outfit. \citet{sarkar2022outfittransformer} proposed the use of the CLS token for predicting the overall compatibility of the outfit. However, after experimentation, we found this to be sub-optimal for the MID task and our architectural approach to perform consistently better. VICTOR is optimized based on two different loss functions. $Y_{OC_{r}}$ - being a regression task - is optimized based on the mean squared error loss function ($L_{MSE}$), while $Y_{MID}$ - being a multi-label classification task - is optimized based on the binary cross entropy ($L_{BCE}$) loss, ignoring the zero padded items.
However, the two loss functions do not necessarily have balanced values. We therefore introduce $\alpha$, a hyper-parameter for weighted combination of the two loss functions as a standard multi-objective optimization practice.
The final loss for VICTOR is calculated as $L = L_{MSE} + L_{BCE} \cdot \alpha$.

In this study, our focus is mainly centered around visual features. However, we also experiment with text in order to be comparable with the current state of the art. For the experiments that also use text, we pass the text descriptions $\mathcal{T} = \{T(g_{1}), T(g_{2}), \dots, T(g_{n})\}$ of outfit $\mathcal{O}$ through a text encoder E\textsubscript{T}($\cdot$) that produces the corresponding vector representations $\mathbf{t}_{g_i}\in\mathbb{R}^{e\times 1}$. $\mathbf{t}_{g_i}$ are concatenated with $\mathbf{v}_{g_i}$ and passed through the transformer decoder D($\cdot$). Thereafter, the following steps are identical with the ones described for the image-only experiments.

\subsection{Fashion-specific language image pre-training (FLIP)}

\begin{figure*}[!ht]
    \centering
    \includegraphics[width=0.8\textwidth]{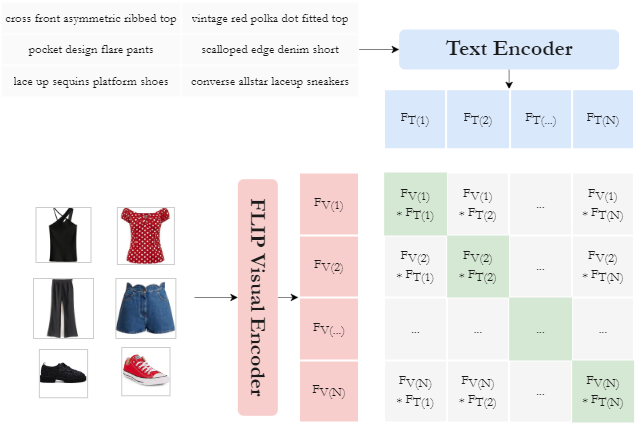}
    \caption{Workflow of fashion language-image pre-training (FLIP). FLIP consists of a visual and a textual encoder that are trained contrastively to predict the correct image-text pair which are placed in the main diagonal. 
    Images and texts are selected with in-batch sampling. 
    }
    \label{fig:FLIP}
\end{figure*}

Analysing the visual compatibility of fashion items requires the use of computer vision neural networks for producing informative representations of said items.
Unlike previous works that have utilized feature extraction from ImageNet-pretrained models or end-to-end fine-tuning, we propose the use of contrastive language-image pre-training for fashion imagery (FLIP). FLIP's workflow is illustrated in Fig. \ref{fig:FLIP} and is following the training procedure proposed by \citet{radford2021learning}.
FLIP consists of one visual E\textsubscript{V}($\cdot$) and one textual E\textsubscript{T}($\cdot$) encoder. Image-text pairs ($I(g_i)$, $T(g_i)$) are passed through their respective encoders and the resulting embeddings are projected onto the same embedding space with the use of two fully connected layers of the same size one for each encoder, as shown below:

\begin{equation}
    \text{F\textsubscript{V}(i)}=\textbf{W\textsubscript{V}}\cdot\text{E\textsubscript{V}}(I(g_i))
\end{equation}

\begin{equation}
    \text{F\textsubscript{T}(i)}=\textbf{W\textsubscript{T}}\cdot\text{E\textsubscript{T}}(T(g_i))
\end{equation}

The dot product between image and text projection embeddings are calculated and the loss function is defined as the mean cross entropy between the predicted and the target image-text pairs, the latter being reflected by the main diagonal. Our rationale for utilizing FLIP is that it balances performance and efficiency. Training computer vision models end-to-end for outfit compatibility can yield a high performance but is a rather resource intensive process. On the other hand, ImageNet-pretrained models do not specialise on fashion imagery and can only produce a general visual representation.
In contrast, the visual encoder of FLIP will learn to produce fashion-specific features. FLIP does not require annotated fashion datasets, which are expensive and time consuming to produce, instead it relies on image-texts pairs of existing fashion products which are easier to attain. Moreover, we may train a single FLIP model, extract the visual features from fashion imagery and textual features from product descriptions and re-use them for numerous experiments on outfit compatibility, such as hyper-parameter tuning and ablation analyses, without requiring e2e fine-tuning. Thus, significantly reducing floating point operations (FLOPs) and by extension computational costs and training time. 

\section{Experimental Setup}
\label{sec:exp}

\subsection{Polyvore Dataset}
\label{sec:polyvore}

The Polyvore dataset is a widely used benchmark dataset for outfit recommendation that was collected by \citet{vasileva2018learning}. The dataset provides 68,306 matching outfits comprising 251,008 unique garments. Each garment comes with multi-modal information including an image, product name, description and associated fashion categories consisting of 14 \textit{types} and 142 categories including \textit{bottoms}: ``skirt'', ``long skirt'', \textit{tops}: ``sweater'', ``turtleneck sweater'', \textit{shoes}: ``boots'', ``flat sandals'' but also \textit{hats}, \textit{jewelry} and other \textit{accessories}. 
For every matching outfit the authors have generated an equal amount of fully incompatible outfits by randomly replacing each garments with items of the same category. 
The dataset comes in two versions that have fixed training, validation and testing splits. 
The first version of Polyvore consists of 106,612, 10,000, 20,000 outfits for training, validation and testing respectively. There are no overlapping outfits between the different splits but garments can overlap between the splits.
The second version, Polyvore-Disjoint, consists of 33,990, 6,000, 30,290 outfits for training, validation and testing but there are no overlapping garments between the splits. 
Each outfit has at least 2, a maximum of 19 and a median value of 5 garments.
As the target variable $T_{OC_b}$, fully compatible outfits have a score of 1 while fully incompatible have 0. 

Outfits with more than 10 garments make up less than 0.5\% of all outfits and could therefore be considered outliers. Moreover, we found that outfits with more than 10 garments often have more than one garment of the same category e.g. two pants or two jackets, which is infeasible. However, we do not filter anything out so as to ensure comparability with previous works. 

\subsection{Polyvore-MISFITs Dataset}
\label{sec:polyvore_misfit}

\begin{figure*}[!ht]
    \centering
    \includegraphics[width=0.75\textwidth]{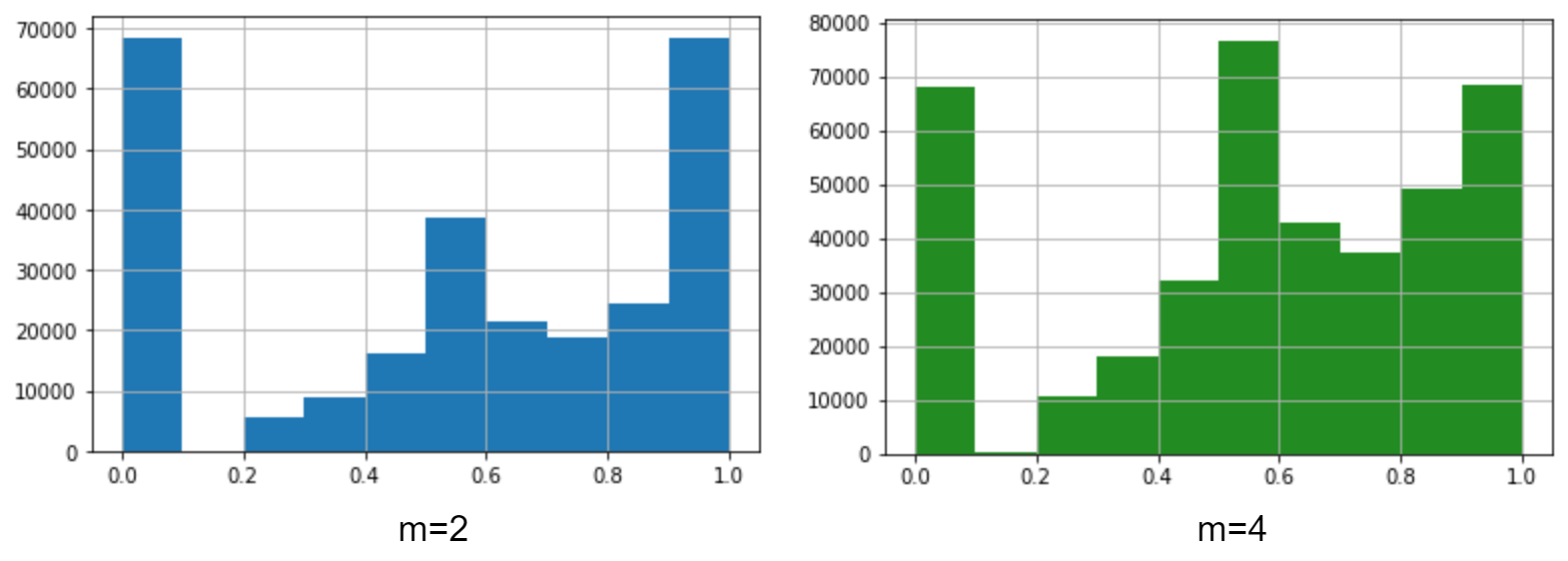}
    \caption{Distribution of compatibility scores for Polyvore-MISFITs with m=2 and m=4.}
    \label{fig:gen_misfits}
\end{figure*}

We apply the MISFIT generation process described in section \ref{sec:gen} on the Polyvore dataset for $m=2$ and $m=4$. $m=2$ creates a balanced dataset between the initial and the generated outfits, with 133,944 MISFITs out of the 270,556 in total which are distributed into 104,498, 9,794, 19,652 for training, validation and testing. This means that 25\% of the dataset consists of fully compatible, 50\% generated MISFITs and 25\% fully incompatible outfits for $m=2$. 
$m=4$ generates 267,888 MISFITs with a total of 404,500 outfits which are split into 315,608, 29,588, 59,304 with 16.88\% of the dataset being fully compatible, 66.22\% generated MISFITs and 16.88\% fully incompatible outfits. The distribution of the compatibility scores for the Polyvore-MISFITs dataset are illustrated in Fig. \ref{fig:gen_misfits}. 

\begin{figure*}[!ht]
    \centering
    \includegraphics[width=0.8\textwidth]{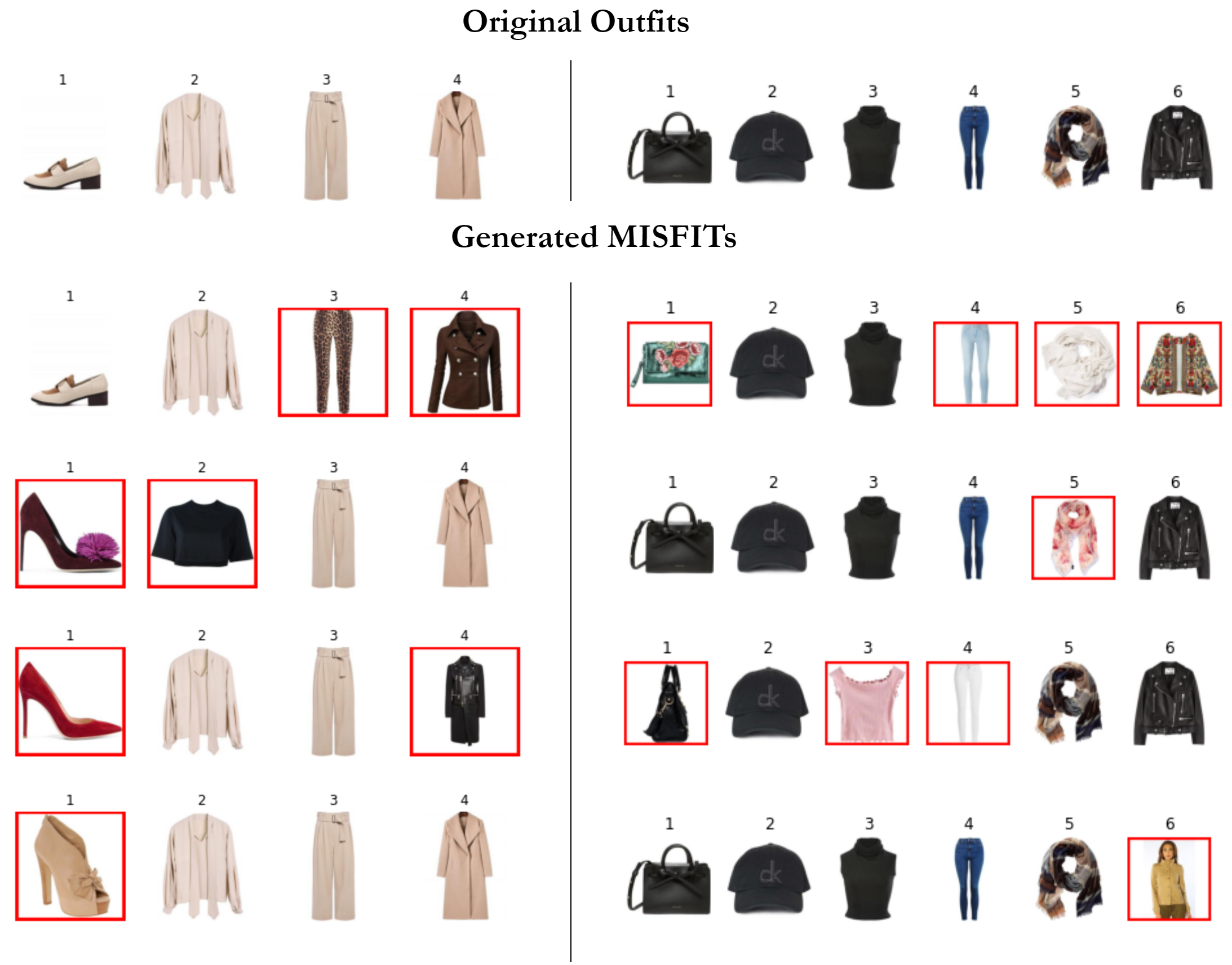}
    \caption{Examples of generated MISFITs from fully compatible outfits. Red frames denote the mismatching items.}
    \label{fig:misfit_gen}
\end{figure*}

Fig. \ref{fig:misfit_gen} presents two indicative examples of generated MISFITs. On top there are two women's outfits of different styles which are annotated as matching. On the left, a classic monochromatic look with a loose fit and on the right a casual look with black pieces and blue jeans. 
The MISFIT generation process has randomly replaced certain garments of the original outfit with items of the same category. For example, the beige pair of wide-fitting pants is replaced with leopard-print leggings (item 3, row 1) and the leather jacket (right outfit) is replaced with a colorful Aztec-pattern jacket (item 6, row 1). These, like most replaced garments, are not matching the aesthetic and style of the initial outfit. Thus, they are correctly categorised as mismatching items. 
However, the generation process is not perfect. For example, the pair of beige loafers is replaced with a beige pair of heels (item 1, row 4, left outfit) which some would not consider it to be mismatching with the rest of the outfit. The replacement process is random, thus, replacement items may match by chance. 
However, since each category contains thousands of items, we expect that random selections will more often than not lead to incompatible combinations. We provide the code\footnote{\url{https://github.com/stevejpapad/Visual-InCompatibility-Transformer}} for generating the Polyvore-MISFITs for reproducibility and in order to encourage further research in the field. 

\subsection{Implementation Details}

We perform an ablation and comparative analysis and in order to distinguish different versions of VICTOR, we denote the training task in square brackets. The proposed multi-tasking learning (MTL) model optimized both for $OC_{r}$ and MID is referred to as VICTOR[MTL].
Furthermore, we define: (1) VICTOR[$OC_{b}$] trained only for binary outfit compatibility, optimized based on the binary cross entropy loss function,
(2) VICTOR[$OC_{r}$] trained only for compatibility as regression optimized based on on the MSE loss function and (3) VICTOR[MID] trained only for mismatching item detection, optimized based on the multi-label binary cross entropy loss function. 
For all versions of VICTOR, we select $L = 8$ transformer layers of $d = 64$ dimensions, $h = 16$ attention heads, a dropout rate of 0.2 and a batch size of 512. 
We train VICTOR[MTL] four times with $\alpha \in [0.2, 0.5, 1, 2]$ and denote different values of $\alpha$ as VICTOR[MTL;$\alpha$].
Wherever required, we also denote the version of Polyvore-MISFITs that was used to train VICTOR as VICTOR[MTL;$\alpha$;$m$].

We use the image-text pairs from the Polyvore-Disjoint dataset for training FLIP since there is no overlap between training, validation and testing sets.
For FLIP's visual encoder \text{E\textsubscript{V}}, we experiment with four models 1) ResNet18 \cite{he2016deep}, 2) EfficientNetV2-B3 \cite{tan2021efficientnetv2}, 3) MLP-Mixer B/16 \cite{tolstikhin2021mlp} and ViT B/32 \cite{dosovitskiy2020image}. 
The aformentioned models are taken from the timm library\footnote{\url{https://github.com/rwightman/pytorch-image-models}} and are initially pre-trained on ImageNet. 
The input image sizes are 224 for all models expect EfficientNetV2-B3 which is 300.
For FLIP's textual encoder \text{E\textsubscript{T}}, we use CLIP's Transformer text encoder and do not fine-tune it any further.
We select a projection layer of 512 and a batch size of 32 for FLIP.

We train both FLIP and VICTOR for 20 epochs with the Adam optimizer and a learning rate scheduler with an initial learning rate of 1e-4 that reduces by a rate of 0.1 at 10 epochs. 

Regarding the evaluation protocol, we follow all previous works that use the area under the roc curve (AUC) as the evaluation metrics for $OC_{b}$. For $OC_{r}$ we report the mean absolute error (MAE) and for MID the binary accuracy and exact match. We use the training, validation and testing sets as provided by the Polyvore dataset in order to ensure fair comparability.
We checkpoint the network's parameters with TOPSIS \cite{hwang1981methods} based on the validation MAE, binary accuracy and exact match.

\section{Results}
\label{sec:results}

\subsection{FLIP and FLOPs}
\label{sec:flip_results}

\begin{table}[!ht]
  \centering
  \caption{Performance of computer vision models fine-tuned with FLIP in terms of the cross entropy loss.}
  \label{tab:flip_flops}
  \begin{tabular}{cc}
  \midrule
     \multicolumn{1}{c}{Model} & 
     \multicolumn{1}{c}{Cross entropy loss 
     ($\downarrow$)
     }
     \\
     \midrule
     ViT B/32 & 1.27 \\          
     ResNet18 & 1.23 \\
     MLP-Mixer B/16 & 1.21 \\     
     EfficientNetV2-B3 & \textbf{1.07} \\     
    \bottomrule
  \end{tabular}
\end{table}

We fine-tune four computer vision neural networks for fashion imagery with the use of fashion language-image pre-training (FLIP). Their performance in terms of the cross entropy loss can be seen in Table \ref{tab:flip_flops}. Lower values of cross entropy loss translates into fewer mistakes when matching the visual and textual projections of actual image-text pairs. However, lower cross entropy loss may not necessarily translate into better performance for VICTOR. 
Our rationale for employing FLIP was to fine-tune the models on fashion imagery while avoiding end-to-end (E2E) fine-tuning for outfit compatibility which can be considerably resource-intensive. 

\begin{table*}
\centering
  \caption{Floating-point operations (FLOPs) of VICTOR when trained with FLIP or end-to-end (E2E) fine-tuning with different computer vision models.}
  
  \label{tab:flops}
  \begin{tabular}{lcccccc}
    \toprule

     \multicolumn{1}{l}{\textbf{Model}} & 
     \multicolumn{1}{c}{\textbf{Model Parameters}} &      
     \multicolumn{1}{c}{\textbf{FLIP}} & 
     \multicolumn{1}{c}{\textbf{VICTOR}} & 
     \multicolumn{1}{c}{\textbf{FLIP + VICTOR }} &
     \multicolumn{1}{c}{\textbf{VICTOR (E2E)}} & 

     \multicolumn{1}{c}{\textbf{\% $\downarrow$}}
     \\
     \midrule
     
     ResNet18 & 1.14E+07 & 5.36E+09 & 1.82E+08 & 5.54E+09 & 4.55E+10 & 87.8 \\
     
     EfficientNetV2-B3 & 1.30E+07 & 6.07E+09 & 1.55E+09 & 7.63E+09 & 6.02E+10 & 87.3 \\
     
     MLP-Mixer B/16 & 5.93E+07 & 7.31E+09 & 4.00E+08 & 7.71E+09 & 2.40E+11 & 96.8 \\
     
     ViT B/32 & 8.76E+07 & 1.56E+10 & 4.00E+08 & 1.60E+10 & 8.26E+10 & 80.62 \\

    \bottomrule
  \end{tabular}
\end{table*}

To measure the efficiency gains of FLIP, we calculate the number of floating point operations (FLOPs) using Facebook's  \textit{fvcore}\footnote{\url{https://github.com/facebookresearch/fvcore}}. 
Table \ref{tab:flops} presents the FLOPs of each computer vision model for a single instance of training. 
We observe that employing FLIP and then utilizing the extracted visual features to train VICTOR reduces the number of FLOPs by an average of 88.14\% compared to E2E training.
Moreover, if we not only consider instance-wise FLOPs but also epoch-wise FLOPs there is an average decrease of up to 94.86\%. This is due to FLIP being trained on the Polyvore-Disjoint dataset (86,624 training+validation instances) - but is then also used for Polyvore - compared to the Polyvore's 202,446 training+validation instances. 
Furthermore, we should also consider the re-usability of FLIP, meaning that a FLIP model can be trained once but its extracted features can then be re-used with no additional cost. In our study, we run 12 experiments per computer vision model, for the ablation study and the tuning of $\alpha$. Compared to using standard E2E training within the same experimental setup, we have actually reduced the number of FLOPs by an impressive average of 98.81\%. 
Utilizing FLIP proved to be significantly more efficient than conventional E2E training for outfit compatibility prediction. 

\begin{table*}[!ht]
\centering
  \caption{Ablation analysis between VICTOR[$OC_{r}$], VICTOR[MID] and VICTOR[MTL] on Polyvore-MISFITs dataset with $m=2$ and $m=4$. 
  For VICTOR[MTL] the comparison between single-modal (text-only or image-only) and multi-modal (text+images) inputs is also presented. The best performing $\alpha=a|b$ based on TOPSIS with \textit{a} for $m=2$ and \textit{b} for $m=4$ is reported.   Textual features are extracted from FLIP's text encoder.
  \underline{Underline} denotes the best performance among image-only models while \textbf{bold} denotes the best overall performance.
  }
  \small
  \label{tab:ablation}
  \begin{tabular}{llcccccccc
  }
    \toprule
     \multicolumn{1}{l}{\textbf{VICTOR}} & 
     \multicolumn{1}{l}{\textbf{FLIP Model}} &  
     \multicolumn{2}{c}{\textbf{MAE ($\downarrow$)}} & 
     \multicolumn{2}{c}{\textbf{Exact Match ($\uparrow$)}} & 
     \multicolumn{2}{c}{\textbf{Accuracy ($\uparrow$)}} 
    &  \multicolumn{2}{c}{\textbf{$OC_{b}$ AUC ($\uparrow$)}}
     \\
     
     \midrule
     
     & & $m$=2 & $m$=4 & $m$=2 & $m$=4 & $m$=2 & $m$=4 & $m$=2 & $m$=4  
    \\
     

     \midrule
     
     \multirow{4}{*}{VICTOR[$OC_{r}$]}
     
     & ResNet18 & 0.254 & 0.221 & - & - & - & - 
     & 0.90 & 0.90
     \\

     & EfficientNetV2-B3 & 0.255 & 0.226 & - & - & - & - 
     & 0.91 & 0.88 
     \\     

     & MLP-Mixer B/16 & 0.255 & 0.229 & - &- &- & - 
     & 0.89 & 0.86 
     \\     
     
     & ViT B/32 & 0.254	& 0.225 & - & - & - & - 
     & \underline{0.92} & \underline{0.92} \\
     
     \midrule
     
     \multirow{4}{*}{VICTOR[MID]}
     & ResNet18 & - & - & 38.30 & 26.29 & 68.64 & 69.44
     & 0.89 & 0.90
     \\

     & EfficientNetV2-B3 & - & - & 38.50 & 27.52 &  71.99 &  70.29
     & 0.91 & 0.90
     \\     

     & MLP-Mixer B/16 & - & - & 36.70 & 27.03 & 72.42 & 70.44
     & 0.90 & 0.90 
     \\     
     & ViT B/32 & - & - & 37.69 & 27.85 & \underline{72.79} &  70.59 
     & 0.91 & 0.90 
    \\ 
     
     \midrule
     
     \multirow{9}{*}{VICTOR[MTL]}
     & ResNet18 ($\alpha=0.2 | 0.2$) & 0.257 & 0.224 & 40.70 & 26.02 & 69.35 & 65.75 
     & 0.90 & 0.90
    \\

     & EfficientNetV2-B3 ($\alpha=0.2 | 1$) & 0.248 & 0.216 & 39.70 & 26.98 & 70.24 & 69.79
     & 0.91 & 0.91 
     \\   
     & MLP-Mixer B/16 ($\alpha=0.2 | 0.2$) & \underline{0.247} & 0.222 & \textbf{\underline{41.55}} & 26.15 &  70.57 & 68.98 
     & \underline{0.92} & 0.91 
     \\
     
     & ViT B/32 ($\alpha=0.2 | 1$) & 0.250 & \underline{0.214} & 40.65 & \underline{27.91} & 70.38 & \underline{70.68}
     & \underline{0.92} & \underline{0.92}
     \\   
     
    \cline{2-10}
    
    & Text ($\alpha=0.2 | 0.2$) & 0.293 & 0.243 & 33.91 & 20.94 & 63.98 & 62.15 & 0.80 & 0.80
    \\

    \cline{2-10}
     & Text + ResNet18 ($\alpha=0.2 | 0.2$) & 0.238 & 0.212 & 39.09 & 27.53  & 72.23 & 70.01 & 0.92 & \textbf{0.93}
    \\

     & Text + EfficientNetV2-B3 ($\alpha=0.2 | 1$) & 0.248 & 0.221 & 38.60  & 23.68 & 70.44 & 66.34 & 0.91 & 0.90
     \\     

     & Text + MLP-Mixer B/16 ($\alpha=0.2 | 0.2$) & 0.238 & 0.212 & 41.09 & \textbf{28.97} & 73.00 & 70.90 & \textbf{0.93} & \textbf{0.93}
     \\
     
     & Text + ViT B/32 ($\alpha=0.2 | 1$) & \textbf{0.230} & \textbf{0.204} & 39.80 & 28.00 & \textbf{73.29} & \textbf{71.28} & \textbf{0.93} & \textbf{0.93}
     \\

    \bottomrule
  \end{tabular}
\end{table*}

\subsection{Ablation Analysis}
\label{sec:ablation}

We perform an ablation analysis comparing the proposed multi-tasking VICTOR[MTL] with its two separate components, VICTOR[$OC_{r}$] and VICTOR[MID]. The results are shown in Table \ref{tab:ablation}. 
For VICTOR we tune the $\alpha$ hyper-parameter - the weight that combines the two loss functions - and report the best performing based on TOPSIS which takes into account \textit{MAE}, exact match and binary accuracy.
VICTOR[$OC_{r}$] is only trained for compatibility prediction as regression and can not detect specific mismatching items. 
Being specialised on $OC_{r}$, it yields an average \textit{MAE} of 0.255 for $m=2$ and 0.225 for $m=4$. VICTOR[MTL] performs marginally better with 0.250 for $m=2$ and 0.219 for $m=4$. 
The overall lowest, hence better, MAE scores are reached by VICTOR[MTL] with MLP-Mixer B/16 and ViT B/32 for $m=2$ and $m=4$ respectively. 

VICTOR[MID] is trained on predicting mismatching items in outfits and yields on average a \textit{binary accuracy} of 71.46\% for $m=2$ and 70.19\% for $m=4$,
closely followed by VICTOR[MTL] which has 70.14\% and 68.8\% respectively.
In terms of the \textit{exact match} evaluation metric, the strictest evaluation metric for the MID task, we observe that VICTOR[MTL] significantly outperform VICTOR[MID] with 40.65\% compared to 37.80\% for $m=2$ while they both perform similarly for $m=2$, with 26.77\% and 27.17\% accordingly.
The overall highest, hence better, exact match scores are reached by VICTOR[MTL] with MLP-Mixer B/16 and ViT B/32 for $m=2$ and $m=4$ respectively.
Regarding binary outfit compatibility prediction ($OC_{b}$), which is evaluated in terms of \textit{AUC}, we observe that VICTOR[MTL], with 0.91/0.91 AUC on average for $m=2$/$m=4$ respectively, slightly outperforming VICTOR[$OC_{r}$]: 0.91/0.89 and MID-only Transformer: 0.90/0.90. ViT B/32 reaches the highest $OC_{b}$ AUC (0.92) for both m=2 and m=4 with either VICTOR[$OC_{r}$] or VICTOR[MTL].

Based on TOPSIS, the overall best performance, among image-only models, is reached by VICTOR[MTL;$\alpha=0.2$;$m=2$] with visual features from MLP-Mixer B/16.
We observe that combining $OC_{r}$ and MID in one model and tuning the hyper-parameter $\alpha$, consistently performs well on both tasks with all computer vision models.
Presumably, by addressing two closely related phenomena and tasks simultaneously and from different perspectives, the multi-tasking VICTOR learns to better recognize compatibility relations among various garments.

Finally, Table \ref{tab:ablation} also illustrates the performance of VICTOR[MTL] with different features as input, namely: text-only, image-only and the combination of text and image. We observe that multi-modality, combining textual and visual features, consistently exceeds the text-only and image-only versions of VICTOR. 
Based on TOPSIS, the overall best performance is reached by VICTOR[MTL;$\alpha=0.2$;$m=2$] with visual features from ViT B/32 and textual features from FLIP's text encoder.

\subsection{Comparative Analysis}

The central focus of this study is the detection of mismatching items in outfits which can be considered a sub-task of visual compatibility. However, to the best of the authors knowledge, no previous works have addressed these tasks and there are no available models to compare VICTOR with.
Instead, we compare the proposed VICTOR[MTL] with numerous state-of-the-art (SotA) models for binary outfit compatibility prediction ($OC_{b}$) which is closely related to $OC_{r}$. 
The current SotA for visual-based $OC_{b}$ on the Polyvore dataset is held by CSA-Net \cite{lin2020fashion} and OutfitTransformer \cite{sarkar2022outfittransformer} with 0.91 AUC. When category information are added the performance of OutfitTransformer increases to 0.92 and when texts are also added it yields 0.93 AUC. 

\begin{table*}[!ht]
  \centering
  \caption{Comparison with the state-of-the-art on binary outfit compatibility prediction ($OC_{b}$) in terms of AUC. For VICTOR, we report the best performing hyper-parameter combination. \underline{Underline} denotes the best performance among image-only models while \textbf{bold} denotes the best overall performance.}
  \label{tab:sota}
  \begin{tabular}{llcc}
    \toprule
     \multicolumn{1}{l}{\textbf{Method}} & 
     \multicolumn{1}{l}{\textbf{Input}} & 
     \multicolumn{1}{c}{\textbf{Polyvore}} & 
     \multicolumn{1}{c}{\textbf{Polyvore-D}}
     \\
     \midrule
     BiLSTM + VSE \cite{han2017learning} & ResNet18 (E2E) + Text & 0.65 & 0.62 \\
     
     GCN (k=1) \cite{li2019coherent} & ResNet18 (E2E) & 0.82 & 0.87 \\
     
     \citet{li2019coherent} & ResNet18 (E2E) & 0.90 & 0.85 \\
     
     SiameseNet \cite{vasileva2018learning} & ResNet18 (E2E) & 0.81 & 0.81 \\
     
     Type-aware \cite{vasileva2018learning} & ResNet18 (E2E) + Text & 0.86 & 0.84 \\
     
     SCE-Net \cite{tan2019learning} & ResNet18 (E2E) + Text & 0.91 & - \\
     
     CSA-Net \cite{lin2020fashion} & ResNet18 (E2E) & 0.91 & 0.87 \\
     
     OutfitTranformer \cite{sarkar2022outfittransformer} & ResNet18 (ImageNet) & 0.82 & - \\     
     
     OutfitTranformer \cite{sarkar2022outfittransformer} & ResNet18 (E2E) & 0.91 & - \\          
     
     OutfitTranformer \cite{sarkar2022outfittransformer} & ResNet18 (E2E) + Text & \textbf{0.93} & 0.88 \\     
     
     \midrule
     
     \multirow{8}{*}{VICTOR[$OC_{b}$]} & ResNet18 (ImageNet) & 0.86 & 0.78 \\

      & EfficientNetV2-B3 (ImageNet) & 0.86 & 0.78 \\
     
      & MLP-Mixer B/16 (ImageNet) & 0.84 & 0.73 \\
     
      & ViT B/32 (ImageNet) & 0.86 & 0.80 \\     
     
     \cline{2-4}
     
      & ResNet18 (FLIP) & 0.90 & 0.85 \\

      & EfficientNetV2-B3 (FLIP) & 0.91 & 0.86 \\

      & MLP-Mixer B/16 (FLIP) & 0.91 & 0.86 \\

      & ViT B/32 (FLIP) & 0.91 & 0.87 \\ 
      
     \midrule

   \multirow{12}{*}{VICTOR[MTL]} & ResNet18 (ImageNet) & 0.86 & 0.77  \\
    
    & EfficientNetV2-B3 (ImageNet) & 0.86 & 0.79 \\
    
    & MLP-Mixer B/16 (ImageNet) & 0.84 & 0.71 \\
    
    & ViT B/32 (ImageNet) & 0.86 & 0.78 \\

     \cline{2-4}

    & ResNet18 (FLIP) & 0.90 & 0.85 \\
    & EfficientNetV2-B3 (FLIP) & 0.91 & 0.87 \\
    & MLP-Mixer B/16 (FLIP) & \underline{0.92} & 0.87 \\
    & ViT B/32 (FLIP) & \underline{0.92} & \underline{0.88} \\
    
     \cline{2-4}

    & ResNet18 + Text (FLIP) & \textbf{0.93} & 0.88  \\
    & EfficientNetV2-B3 + Text (FLIP) & 0.91 & 0.88  \\
    & MLP-Mixer B/16 + Text (FLIP) & \textbf{0.93} &  0.89 \\
    & ViT B/32 + Text (FLIP) & \textbf{0.93} & \textbf{0.90} \\
    
    \bottomrule
  \end{tabular}
\end{table*}

Comparing the models that use pre-trained visual features on ImageNet, we observe that OutfitTransformer w/ ResNet18 (ImageNet) yields 0.82 AUC on Polyvore while our VICTOR[$OC_{b}$] w/ ResNet18 (ImageNet) outperforms it with 0.86 AUC. 
VICTOR[$OC_{b}$] exhibit a similar performance with the other three computer vision models, with an average AUC of 0.86. 
Furthermore, when employing $\mathbf{v}_{g_i}$ from FLIP, VICTOR[$OC_{b}$] w/ ResNet18 (FLIP) improves to 0.9 AUC similarly with all other computer vision models; that display an average AUC of 0.91 for Polyvore and 0.86 on Polyvore-D.
The proposed VICTOR[MTL;$\alpha=0.2$;$m=2$] further improves upon VICTOR[$OC_{b}$] with MLP-Mixer B/16 and ViT B/32 FLIP models. 
This slight improvement can be attributed to $Y_{OCr}$ forcing VICTOR[MTL] to learn deeper and more complicated relations compared to the simple $OC_{b}$-based model.
Notably, VICTOR[MTL] w/ ResNet18 (FLIP) performs at the same level as the SotA while being significantly faster and less resource-intensive to train; requiring 94.8\% fewer FLOPs.
VICTOR[MTL] w/ MLP-Mixer B/16 (FLIP) or ViT B/32 (FLIP) surpasses the vision-based SotA with 0.92 AUC on Polyvore while
VICTOR[MTL] with ViT/B32 (FLIP) surpasses the SotA on Polyvore-D with 0.88 AUC.

When textual features are added, we observe that VICTOR[MTL] with ResNet18 exhibits the same AUC (0.93) as OutfitTransformer on the Polyvore dataset. 
Moreover, VICTOR[MTL] with either MLP-Mixer B/16 and ViT B/32 surpass OutfitTransformer on the Polyvore-D dataset with 0.89 and 0.90 AUC scores respectively without requiring end-to-end fine-tuning. 
Therefore, the multi-modal VICTOR[MTL] has defined a new SotA score on Polyvore-D - the more challenging version of Polyvore - while maintaining very high efficiency.

\subsection{Qualitative Analysis}

\begin{figure*}[!ht]
\begin{subfigure}{\textwidth}
  \centering
  \includegraphics[width=\linewidth]{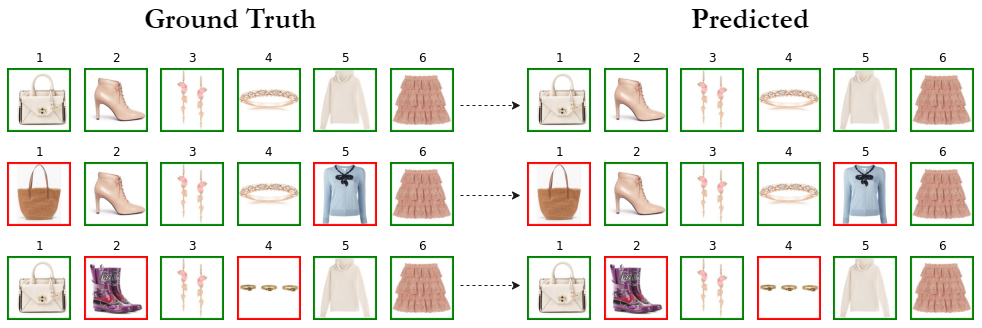}  
  \caption{}
  \label{fig:sub-first}
\end{subfigure}

\begin{subfigure}{\textwidth}
  \centering
  \includegraphics[width=\linewidth]{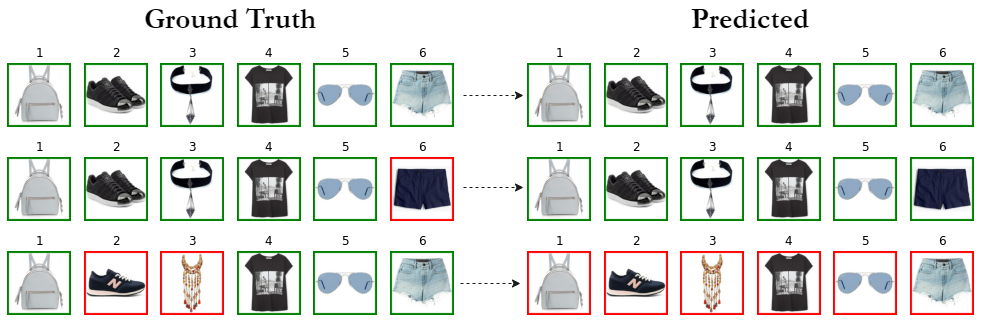}  
  \caption{}
  \label{fig:sub-fourth}
\end{subfigure}

\caption{Inference examples from VICTOR on fully compatible outfits and their generated partially mismatching versions. Green frames denotes compatible items while red frames denote incompatible items.}
\label{fig:inference}
\end{figure*}

Fig.\ref{fig:inference} illustrates two inference examples from VICTOR[MTL;$\alpha=0.2$;$m=2$] with MLP-Mixer B/16 since it exhibited the highest exact match score. 
We use samples from the Polyvore MISFITs $m=2$ thus there are three fully compatible outfits and for each, there are two generated outfits containing at least one incompatible item.
We observe that VICTOR is capable of correctly identifying the fully compatible outfits in 
both cases (row=1 of each outfit). 
There are also cases that correctly identifies all mismatching items,such as row 2 and 3 of Fig. \ref{fig:sub-first}. VICTOR has presumably learned to ``understand'' which styles and colors of different garments can be matched together.

Expectedly, there are also some mistaken predictions. Row 3 of Fig. \ref{fig:sub-fourth} the whole outfit is predicted to be incompatible while $4/6$ items are annotated as compatible.
On the other hand, row 2 of Fig.\ref{fig:sub-fourth} although the pair of dark navy shorts is annotated as mismatching with the rest of the outfit, some would consider this to be a mistaken annotation since grey, black and navy are often paired together. VICTOR seems to have generalized well enough so as to ignore the few rare cases of mistaken annotations. 

One general challenge for the task of visual compatibility is that there is always the element of subjectivity. Moreover, what is considered compatible differs from culture to culture and is time dependent; since fashion trends are in constant flux.
In our case, the ground truth compatible outfits reflect the subjective opinions and biases of fashion stylists from Polyvore, creating a data-driven bias in our models.  
Despite this caveat, overall, VICTOR seems to produce reasonable predictions and we believe that a larger and more diverse dataset would further improve its performance. 

\begin{figure*}[!ht]
    \centering
    \includegraphics[width=0.6\textwidth]{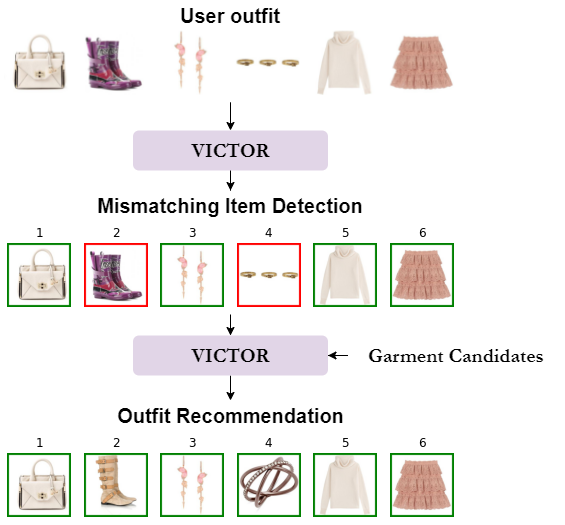}
    \caption{Example of VICTOR detecting the mismatching items in an outfit and recommending more compatible alternatives. Green frames denotes compatible items while red frames denote incompatible items.}
    \label{fig:rec_outfit}
\end{figure*}

Finally, VICTOR does not only predict the mismatching items in an outfit but has also learned to predict the overall compatibility of an outfit. As a result, it can also be used for outfit recommendation.
Fig. \ref{fig:rec_outfit} illustrates an example where VICTOR detects two mismatching items in an outfit and given a set of candidate garments - all garments from the same category as the mismatching items - selects the better suited alternatives, resulting in a more cohesive and aesthetically pleasing outfit.

\section{Conclusions}
\label{sec:conclusions}

In this study we define two new sub-tasks within the general task of visual compatibility prediction, namely compatibility prediction as regression ($OC_r$) and mismatching item detection (MID) and examine both in the Fashion domain. 
We use the Polyvore outfits dataset to generate partially mismatching outfits (MISFITs) and create the Polyvore-MISFITs dataset where we perform a series of ablative and comparative experiments.
We propose a multi-tasking Transformer-based architecture, named VICTOR, and utilize visual features from multiple computer vision neural networks fine-tuned with \textit{fashion-specific contrastive language-image pre-training} (FLIP). 

The ablation study showed that addressing both tasks ($OC_r$ and MID) in a single architecture performs better than single-task models and that multi-modality, combining both textual and visual features from FLIP, outperforms single-modality models.
Furthermore, in the comparative analysis, VICTOR outperformed state-of-the-art models by 4.87\% in terms of AUC on the Polyvore dataset when using visual features extracted from ImageNet-pretrained models with no additional computational cost.
By utilizing features from FLIP, VICTOR was not only capable of competing and even surpassing state-of-the-art methods on Polyvore datasets, but also to reduce instance-wise floating point operations by 88\%.

One limitation of the current study is that when generating the Polyvore-MISFITs dataset, we use random alternative sampling. 
More intricate methods could theoretically be implemented, that take into account the rate of similarity between the ground truth and the selected mismatching garment. However, it is difficult to select the appropriate threshold of similarity without input from professional stylists.
Selecting too similar items - e.g a black pair of dress shoes with another - would not result in actual mismatching outfits but selecting too dissimilar items - e.g the dress shoes with a pair of snow boots - would lead to numerous easy-to-predict MISFITs and as a result, VICTOR would not have learned to discern more subtle and useful cases of visual incompatibility.
A second issue is that VICTOR has been trained on images from Polyvore dataset which depicts individual garments in a white background. This may limit its application to real-world fashion images as worn by people. However, VICTOR could very easily be integrated in a full system, similar to \cite{papadopoulos2022attentive}, that applies garment detection to real world fashion imagery and then extract the visual features of individual garments; given that the garments are fully or mostly visible.

Our focus is centered around general visual compatibility in fashion. 
By relying on the Polyvore dataset VICTOR has learned to reflect the subjective opinions and biases of fashion stylists from Polyvore. 
It would be interesting for future works to re-create similar architectures that also take personalization into account \cite{zhan2021pan}.
Finally, the proposed VICTOR and FLIP fine-tuning are not limited to applications within the Fashion domain. Future works could experiment with other visually-driven domains such as exterior and interior architecture design \cite{aggarwal2018learning}. 

\bibliographystyle{unsrtnat}
\bibliography{template}  






\end{document}